\documentclass[10pt,twocolumn,letterpaper]{article}
\usepackage{cvpr}              
\usepackage{multirow}
\usepackage{CJKutf8}
\newcommand{\zh}[1]{\begin{CJK*}{UTF8}{gbsn}#1\end{CJK*}}
\definecolor{cvprblue}{rgb}{0.21,0.49,0.74}
\usepackage[pagebackref,breaklinks,colorlinks,allcolors=cvprblue]{hyperref}

\def\paperID{72} 
\def\confName{CVPR}
\def\confYear{2026}

\title{Zero-Shot Chinese Character Recognition via Global-Local Dual-Branch Alignment and Hierarchical Inference}

\author{
Wei Cao$^{1}$ \quad Hao Xu$^{1}$ \quad Xiaolei Diao$^{2}$\thanks{Corresponding author.} \\ 
$^{1}$Jilin University, China \\
$^{2}$University College London, UK \\
{\tt\small caowei23@mails.jlu.edu.cn, xuhao@jlu.edu.cn, xiaolei.diao@ucl.ac.uk}
}

\begin{document}
\maketitle
\begin{abstract}
Chinese character categories are extremely large, and unseen characters frequently arise in open-world scenarios, making zero-shot Chinese character recognition an important yet challenging problem. Existing IDS-based retrieval methods usually encode a character image and its ideographic description sequence into a single global vector for matching. Although efficient, such holistic alignment often under-models local component differences. Moreover, directly introducing patch-token level fine-grained interaction suffers from both the noise of structural operators in IDS and the high cost of full-candidate retrieval.To address these issues, we propose a Global-Local Hierarchical Perception Network (GL-HPN), which jointly learns global and local representations of character images and IDS sequences within a unified cross-modal alignment framework. The global branch supports efficient coarse recall, while the local branch improves component-level discrimination through patch-token interaction. We further introduce a structure filtering mask to suppress structurally meaningful but visually non-entity IDS operators in local similarity aggregation. On top of this, we design a coarse-to-fine hierarchical inference strategy that performs global retrieval over the full candidate set and local reranking only on Top-$K$ candidates, followed by parameter-free multiplicative fusion of normalized posterior scores. Experimental results show that GL-HPN achieves competitive performance across multiple zero-shot splits, performs especially well under low-resource settings, and substantially reduces the inference cost of large-scale candidate retrieval.
\end{abstract}    
\section{Introduction}

Chinese character recognition is of great importance in applications such as ancient document digitization, variant character normalization, and open-world document understanding. Unlike the closed-set setting, real-world scenarios often contain a large number of rare characters, variant characters, and newly coined characters that never appear during training, making zero-shot Chinese character recognition a challenging research problem. Although the number of Chinese character categories is extremely large, the set of radicals and spatial composition rules on which they are built is relatively limited\cite{ijcai2023p73}. Many unseen characters can therefore be regarded as recombinations of known radicals under different structural relations. As a result, if a model can learn transferable component representations and composition patterns, it has the potential to generalize effectively to unseen characters\cite{diao2023toward}.

Existing zero-shot Chinese character recognition methods can be broadly divided into two categories. Predictive methods formulate recognition as the explicit prediction of radicals, structures, or stroke sequences, followed by rule-based parsing or lexicon matching to recover the final character. These methods are structurally interpretable, but usually suffer from error accumulation and long inference chains. In contrast, retrieval-based methods cast recognition as a cross-modal matching problem between character images and structural descriptions. Inspired by CLIP-style contrastive learning frameworks, recent methods such as CCR-CLIP\cite{yu2023chinese}, FT-CLIP\cite{hong2025improving}align character images with ideographic description sequence (IDS) representations and have achieved substantial progress in zero-shot Chinese character recognition.

Despite their effectiveness, most existing retrieval-based methods adopt a holistic alignment paradigm, where character images and IDS sequences are encoded into single global vectors and matched via global similarity. While efficient, this paradigm tends to weaken the modeling of local component differences and fine-grained compositional variations\cite{asokan2025finelip}, leading to unstable discrimination in scenarios dominated by local substitutions, subtle structural differences, or visually similar components.As highlighted in recent visual classification studies, obtaining robust and highly discriminative vision feature representations is crucial for overall model performance when dealing with such subtle visual patterns \cite{shi2025competitive}. Furthermore, directly introducing patch-token level fine-grained interaction over the full candidate space leads to two additional challenges. First, structural operators in IDS do not correspond to concrete visual entities and may therefore introduce invalid matching noise during local alignment. Second, the cost of fine-grained interaction grows rapidly with the number of candidate categories, limiting its practicality in large-scale retrieval settings.

To address these issues, we propose a \emph{Global-Local Hierarchical Perception Network} (GL-HPN) for zero-shot Chinese character recognition. Within a unified vision-language alignment framework, GL-HPN jointly models global and local representations of character images and IDS sequences. The global branch learns stable holistic structural alignment for efficient recall over large candidate sets, while the local branch explicitly models the correspondence between image regions and radical tokens through patch-token interaction, thereby enhancing component-level discrimination. Considering that structural operators in IDS lack direct visual correspondence, we further introduce a \emph{structure filtering mask} to suppress non-entity tokens during local similarity aggregation and reduce invalid responses. During inference, we design a coarse-to-fine hierarchical inference strategy that first performs global ranking over the full candidate set and then conducts local reranking only on Top-$K$ candidates. Finally, global and local evidence are fused through multiplicative posterior fusion, improving recognition performance without significantly increasing inference cost.

Our main contributions are summarized as follows:
\begin{enumerate}
    \item We propose a global-local dual-branch cross-modal alignment framework for zero-shot Chinese character recognition, balancing holistic structural modeling and local component-level discrimination.
    \item We introduce a structure filtering mask to explicitly suppress the noise caused by IDS structural operators in local fine-grained matching, improving the stability of local alignment.
    \item We design a coarse-to-fine hierarchical inference strategy that restricts expensive local interaction to a small set of high-confidence candidates, achieving a favorable trade-off between recognition accuracy and inference efficiency.
\end{enumerate}
\section{Related Work}

\subsection{Zero-Shot Chinese Character Recognition}

Zero-shot Chinese character recognition for unseen characters has attracted increasing attention in recent years. Existing methods can be broadly categorized into predictive methods and retrieval-based methods. Predictive methods formulate recognition as the explicit prediction of discrete symbol sequences, such as radicals, structures, or strokes, and then recover the final character through rule-based parsing, lexicon matching, or reasoning modules. Representative works include DenseRAN\cite{wang2018denseran}, STAR\cite{zeng2022star}, RSST\cite{yu2024chinese}, RZCR\cite{ijcai2023p73}. These methods can effectively exploit the compositional regularities of Chinese characters and are often highly interpretable, but they usually suffer from cumulative sequence prediction errors, complex hierarchical decomposition pipelines, and the overhead of post-processing or reasoning modules.

In contrast, retrieval-based methods do not explicitly generate discrete sequences. Instead, they formulate zero-shot recognition as similarity matching in a continuous embedding space. Such methods typically treat structural descriptions of characters as textual priors and learn cross-modal consistency between character images and structural descriptions, enabling retrieval and classification over a candidate set. Inspired by CLIP-style frameworks, CCR-CLIP\cite{yu2023chinese}, FT-CLIP\cite{hong2025improving}, and GRSTR\cite{dong2026graph} have achieved notable progress in zero-shot Chinese character recognition. CCR-CLIP directly aligns character images with ideographic description sequence (IDS) representations. FT-CLIP introduces formation trees and a dedicated tree encoder to strengthen hierarchical structural modeling. GRSTR further exploits graph structures and direction-aware positional encoding to improve the modeling of spatial relations. Overall, retrieval-based methods enjoy simple inference pipelines, good scalability, and natural suitability for large-scale candidate retrieval. However, most existing methods still adopt a holistic alignment paradigm and therefore struggle to adequately model local component differences and fine-grained compositional changes.

\subsection{Contrastive Learning and Fine-Grained Cross-Modal Alignment}

Contrastive learning aims to learn discriminative representations by pulling positive pairs closer and pushing negative pairs apart in the embedding space, and has become a fundamental paradigm for cross-modal representation learning. CLIP\cite{radford2021learning} adopts a dual-encoder architecture to project images and text into a shared embedding space and uses a symmetric image-text contrastive objective to achieve global semantic alignment. Owing to its strong scalability and zero-shot transfer ability, CLIP has been widely used in retrieval-based Chinese character recognition.


However, global-to-global alignment in CLIP compresses both image and text into a single vector, which may weaken the modeling of fine-grained semantic correspondence. FILIP \cite{yao2021filip} addresses this issue by introducing local-to-local interaction on top of CLIP, constructing finer-grained cross-modal matching scores through patch-token level maximum-response aggregation, thereby improving local correspondence modeling between images and text. Our method is technically inspired by this idea. Nevertheless, unlike generic vision-language scenarios, the textual side of zero-shot Chinese character recognition uses IDS, which contains both radical tokens and structural operators. Bridging the semantic gap in such cross-modal alignment is notoriously challenging. Previous studies highlight that representing textual semantics accurately requires explicit awareness of the corresponding visual properties \cite{giunchiglia2023semantics}, and aligning abstract textual hierarchies without concrete visual semantics often leads to suboptimal feature mappings \cite{diaobuilding}. Since structural operators and radical tokens differ significantly in terms of visual correspondence, it is necessary to explicitly account for the local alignment noise introduced by structural operators.

\subsection{Chinese Character Structural Priors and IDS Representation}

\begin{figure}[t]
    \centering
    \includegraphics[width=0.85\columnwidth]{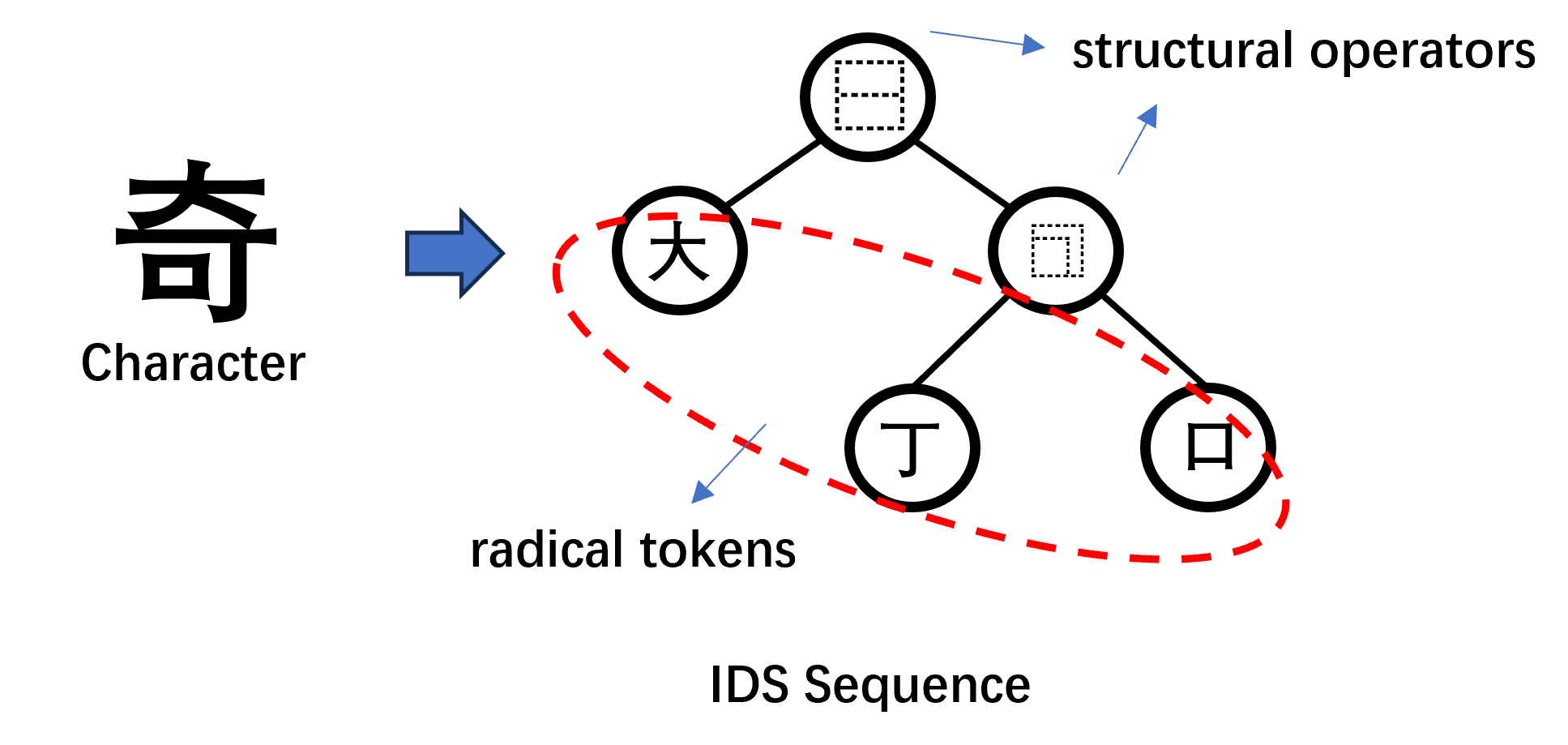}
    \caption{Example of IDS decomposition with radical tokens and structural operators.}
    \label{fig:IDS}
\end{figure}

Chinese characters exhibit strong compositional structure. A large number of characters can be formed from a relatively limited set of radicals and their spatial relations\cite{chen2021benchmarking}. The \emph{Ideographic Description Sequence} (IDS), defined in the Unicode standard\cite{unicode2023unicode}, provides a linear symbolic form that explicitly encodes the component composition and structural relations of Chinese characters, and has therefore been widely adopted in zero-shot Chinese character recognition. As illustrated in Fig.~\ref{fig:IDS} using the character \zh{奇} as an example, an IDS sequence typically consists of two kinds of symbols: radical tokens representing concrete components and structural operators describing spatial composition relations.

Existing methods typically encode a complete IDS sequence into a global representation to achieve cross-modal matching with character images. While this holistic paradigm incorporates structural priors, it often fails to capture fine-grained component variations. Directly introducing patch-token level interaction, however, faces a critical dilemma: structural operators in IDS lack explicit visual correspondence, producing invalid local responses that interfere with the discriminative contribution of radical tokens. Considering that the preservation of character semantics can benefit from anchoring fine-grained alignment on concrete visual entities, we propose the Global-Local Hierarchical Perception Network (GL-HPN). Our method effectively mitigates alignment noise and computational overhead through a structure filtering mask and a coarse-to-fine hierarchical inference strategy.
\section{Method}

\subsection{Overview}

We propose a \emph{Global-Local Hierarchical Perception Network} (GL-HPN) for zero-shot Chinese character recognition. Within a unified vision-language alignment framework, GL-HPN explicitly decouples holistic structural modeling from fine-grained component interaction, and jointly learns global and local representations of character images and ideographic description sequences (IDS). The global branch captures stable sequence-level structural information and supports efficient candidate recall, while the local branch models patch-token correspondence between image regions and radical tokens, thereby improving component-level discrimination.

\begin{figure*}[t]
    \centering
    \includegraphics[width=0.9\textwidth]{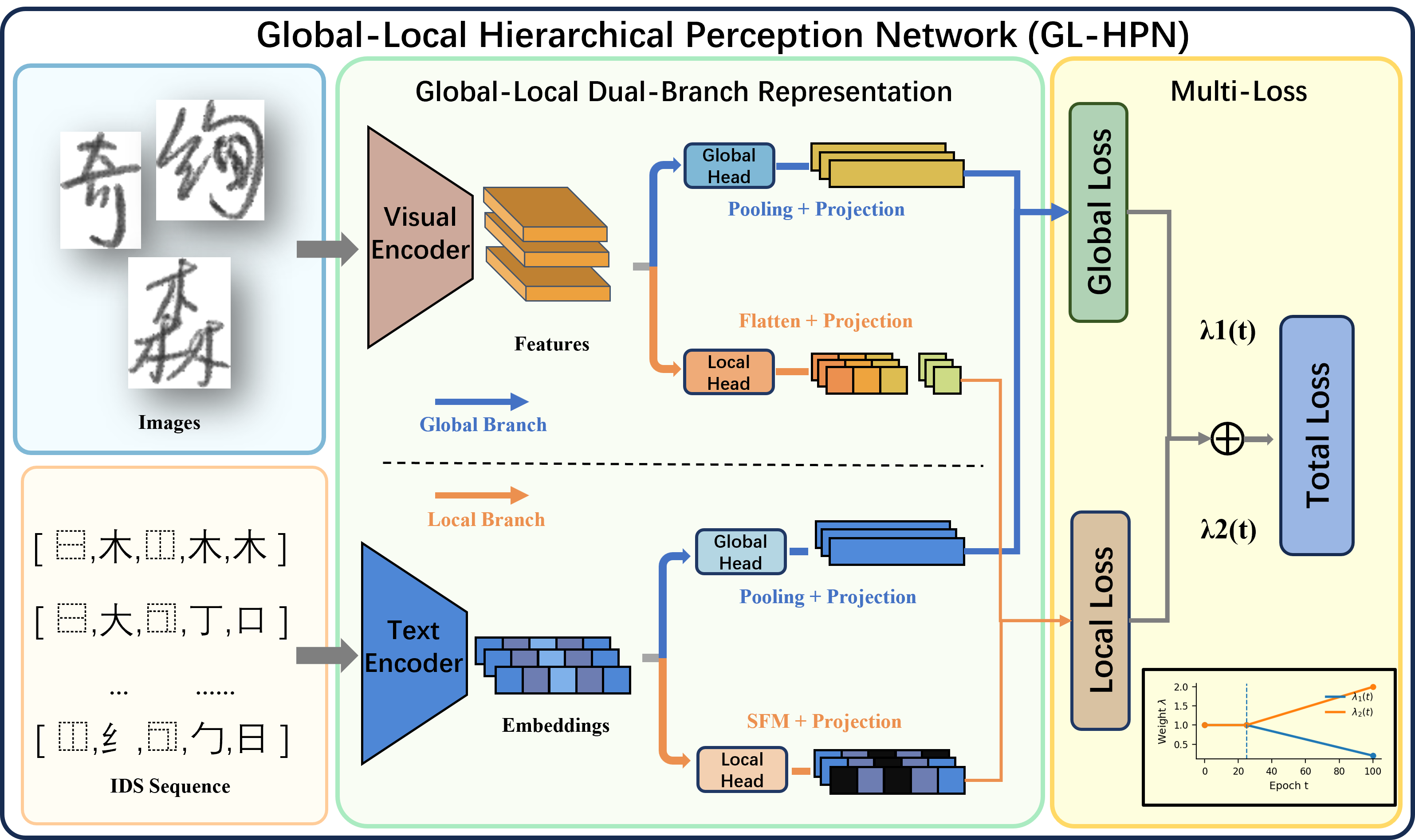}
    \caption{Overview of the proposed GL-HPN. The model learns decoupled global and local representations for both character images and IDS sequences. The global branch performs holistic structural alignment, while the local branch models fine-grained patch-token correspondence under the structure filtering mask.}
    \label{fig:framework}
\end{figure*}

Given a character image $I$ and its IDS sequence $\mathcal{T}$, GL-HPN extracts global and local representations from both modalities within a unified vision-language alignment framework. During training, the model jointly optimizes global and local contrastive objectives with a curriculum-style weighting schedule. During inference, it first performs global candidate recall and then applies local reranking over the Top-$K$ candidates, followed by parameter-free multiplicative posterior fusion. As illustrated in Fig.~\ref{fig:framework}, the overall framework follows a hierarchical recognition pipeline of global structural alignment and local component discrimination.

\subsection{Global-Local Dual-Branch Representation}

\subsubsection{Visual Representation}

Given an input character image $I$, the visual encoder outputs a 2D feature map
\begin{equation}
X = VE(I), \quad X \in \mathbb{R}^{h \times w \times d}
\end{equation}
where $h$, $w$, and $d$ denote the spatial height, width, and channel dimension, respectively. Different from standard image classification settings, the final global pooling and classification layer are removed to preserve spatial layout information.

Based on $X$, we construct a \emph{Global Stream} and a \emph{Local Stream}. For the global branch, global average pooling is applied to obtain an image-level representation
\begin{equation}
z = \mathrm{GAP}(X)
\end{equation}
which is projected into the shared embedding space by
\begin{equation}
v_{\mathrm{global}} = g_{\phi_g}(z)
\end{equation}

For the local branch, a local projection head is applied to the feature map
\begin{equation}
Z = g_{\phi_l}(X)
\end{equation}
and the projected feature map is flattened into a patch sequence
\begin{equation}
V_{\mathrm{local}} = \{v_1, v_2, \ldots, v_{N_p}\}, \quad N_p = h \times w
\end{equation}
Since the global and local heads are decoupled, the two streams can specialize in holistic structural alignment and local component alignment, respectively.

\subsubsection{Textual Representation}

For the IDS sequence of a character, we denote
\begin{equation}
\mathcal{T} = \{\tau_1, \tau_2, \ldots, \tau_M\}
\end{equation}
where $M$ is the sequence length. IDS tokens include radical tokens corresponding to concrete visual components and structural operators describing spatial composition relations.

A Transformer-based text encoder produces token-level features
\begin{equation}
H = TE(\mathcal{T}) = \{h_1, h_2, \ldots, h_M\}
\end{equation}

For the global branch, all token features are averaged to obtain a sequence-level representation
\begin{equation}
u = \frac{1}{M}\sum_{i=1}^{M} h_i
\end{equation}
which is projected into the shared embedding space by
\begin{equation}
t_{\mathrm{global}} = g_{\psi_g}(u)
\end{equation}
Since structural operators participate in contextual encoding and sequence-level aggregation, $t_{\mathrm{global}}$ captures both component information and structural relations.

For the local branch, each token feature is projected to form a token-level local representation sequence
\begin{equation}
T_{\mathrm{local}} = \{t_1^{\mathrm{local}}, t_2^{\mathrm{local}}, \ldots, t_M^{\mathrm{local}}\}
\end{equation}
where
\begin{equation}
t_i^{\mathrm{local}} = g_{\psi_l}(h_i), \quad i=1,2,\ldots,M
\end{equation}

\subsection{Structure Filtering and Multi-Loss Training}

IDS contains radical tokens and structural operators. While radical tokens usually correspond to concrete local regions in the image, structural operators mainly describe spatial composition and do not directly correspond to visual entities. If structural operators are included in patch-token local alignment, they may introduce scattered and non-informative responses. To address this issue, we introduce a \emph{Structure Filtering Mask} (SFM):
\begin{equation}
m_i =
\begin{cases}
1, & \tau_i \in \mathcal{C} \\
0, & \tau_i \in \mathcal{S}
\end{cases}
\end{equation}
where $\mathcal{C}$ denotes the set of radical tokens and $\mathcal{S}$ denotes the set of structural operators. The mask is applied only in the local branch, while structural operators are retained in the global branch to preserve holistic structural information.

Based on the global and local streams, GL-HPN is trained with a multi-loss objective consisting of a \emph{Global Loss}, a \emph{Local Loss}, and a weighted \emph{Total Loss}. Unless otherwise specified, all global and local features are $\ell_2$-normalized before similarity computation, and $\langle \cdot,\cdot \rangle$ denotes cosine similarity.

\subsubsection{Global Loss}

For the global branch, we adopt a CLIP-style contrastive objective to align holistic image and IDS representations. Let the global visual and textual features in a batch be $\{v_{\mathrm{global}}^{(i)}\}_{i=1}^{B}$ and $\{t_{\mathrm{global}}^{(i)}\}_{i=1}^{B}$, respectively. The image-to-text global alignment loss is defined as
\begin{equation}
L_{\mathrm{global}}^{I\rightarrow T}
=
-\frac{1}{B}\sum_{i=1}^{B}
\log
\frac{
\exp\left(\langle v_{\mathrm{global}}^{(i)}, t_{\mathrm{global}}^{(i)} \rangle / \tau_g\right)
}{
\sum_{j=1}^{B}
\exp\left(\langle v_{\mathrm{global}}^{(i)}, t_{\mathrm{global}}^{(j)} \rangle / \tau_g\right)
}
\end{equation}
where $\tau_g$ is the temperature parameter of the global branch. The text-to-image loss is defined symmetrically:
\begin{equation}
L_{\mathrm{global}}^{T\rightarrow I}
=
-\frac{1}{B}\sum_{i=1}^{B}
\log
\frac{
\exp\left(\langle t_{\mathrm{global}}^{(i)}, v_{\mathrm{global}}^{(i)} \rangle / \tau_g\right)
}{
\sum_{j=1}^{B}
\exp\left(\langle t_{\mathrm{global}}^{(i)}, v_{\mathrm{global}}^{(j)} \rangle / \tau_g\right)
}
\end{equation}
The final global loss is obtained by averaging the two directions:
\begin{equation}
L_{\mathrm{global}}
=
\frac{1}{2}
\left(
L_{\mathrm{global}}^{I\rightarrow T}
+
L_{\mathrm{global}}^{T\rightarrow I}
\right)
\end{equation}

\subsubsection{Local Loss}

For the local branch, we adopt FILIP-style maximum-response aggregation to explicitly model fine-grained correspondence between image patches and textual tokens. Given local visual features
\begin{equation}
V_{\mathrm{local}}=\{v_n\}_{n=1}^{N_p}
\end{equation}
and local textual features
\begin{equation}
T_{\mathrm{local}}=\{t_i^{\mathrm{local}}\}_{i=1}^{M},
\end{equation}
the token-driven local similarity is defined as
\begin{equation}
S_{I2T}(V_{\mathrm{local}},T_{\mathrm{local}}) =
\frac{1}{\sum_{i=1}^{M} m_i}
\sum_{i=1}^{M} m_i \max_{n=1,\ldots,N_p} \left\langle v_n, t_i^{\mathrm{local}} \right\rangle
\end{equation}
This aggregation measures, for each valid textual token, the strongest response over all visual patches, and then averages over radical tokens only.

Similarly, the patch-driven local similarity is defined as
\begin{equation}
S_{T2I}(V_{\mathrm{local}},T_{\mathrm{local}}) =
\frac{1}{N_p}
\sum_{n=1}^{N_p} \max_{i:m_i=1} \left\langle v_n, t_i^{\mathrm{local}} \right\rangle
\end{equation}
This aggregation measures, for each visual patch, the strongest response among valid textual tokens and then averages across all patches.

Based on these two local matching scores, we construct bidirectional local contrastive objectives over the batch. Let the local visual and textual features of the $i$-th sample be denoted by $V_{\mathrm{local}}^{(i)}$ and $T_{\mathrm{local}}^{(i)}$, respectively. The image-to-text local alignment loss is defined as
\begin{equation}
L_{\mathrm{local}}^{I\rightarrow T}
=
-\frac{1}{B}\sum_{i=1}^{B}
\log
\frac{
\exp\left(S_{I2T}\!\left(V_{\mathrm{local}}^{(i)},T_{\mathrm{local}}^{(i)}\right)/\tau_l\right)
}{
\sum_{j=1}^{B}
\exp\left(S_{I2T}\!\left(V_{\mathrm{local}}^{(i)},T_{\mathrm{local}}^{(j)}\right)/\tau_l\right)
}
\end{equation}
where $\tau_l$ is the temperature parameter of the local branch. Symmetrically, the text-to-image local alignment loss is defined as
\begin{equation}
L_{\mathrm{local}}^{T\rightarrow I}
=
-\frac{1}{B}\sum_{i=1}^{B}
\log
\frac{
\exp\left(S_{T2I}\!\left(V_{\mathrm{local}}^{(i)},T_{\mathrm{local}}^{(i)}\right)/\tau_l\right)
}{
\sum_{j=1}^{B}
\exp\left(S_{T2I}\!\left(V_{\mathrm{local}}^{(j)},T_{\mathrm{local}}^{(i)}\right)/\tau_l\right)
}
\end{equation}
The final local loss is obtained by averaging the two directions:
\begin{equation}
L_{\mathrm{local}}
=
\frac{1}{2}
\left(
L_{\mathrm{local}}^{I\rightarrow T}
+
L_{\mathrm{local}}^{T\rightarrow I}
\right)
\end{equation}

\subsubsection{Total Loss}

The overall training objective is
\begin{equation}
L_{\mathrm{total}} = \lambda_1(t)L_{\mathrm{global}} + \lambda_2(t)L_{\mathrm{local}}
\end{equation}
where $\lambda_1(t)$ and $\lambda_2(t)$ are the epoch-dependent weights of the global and local losses.

To stabilize joint optimization, we adopt a curriculum-style weighting schedule. Let $T$ denote the total number of training epochs and $T_w$ the warm-up stage. The two weights are defined as
\begin{equation}
\lambda_1(t)=
\begin{cases}
1.0, & t<T_w\\
1.0-\alpha \dfrac{t-T_w}{T-T_w}, & t\ge T_w
\end{cases}
\end{equation}
\begin{equation}
\lambda_2(t)=
\begin{cases}
1.0, & t<T_w\\
1.0+\beta \dfrac{t-T_w}{T-T_w}, & t\ge T_w
\end{cases}
\end{equation}
In our implementation, $T_w$ is set to 25\% of the total training epochs, with $\alpha=0.8$ and $\beta=1.0$. This schedule first emphasizes stable global alignment and then progressively strengthens local fine-grained discrimination.

\subsection{Coarse-to-Fine Hierarchical Inference}

During inference, directly performing local fine-grained interaction over the full candidate set is computationally expensive. To balance recognition performance and inference efficiency, we adopt a coarse-to-fine hierarchical inference strategy.
For notational convenience, we use $v_{\mathrm{global}}(I)$ and $V_{\mathrm{local}}(I)$ to denote the global and local visual representations extracted from input image $I$. Each candidate category $y$ is associated with an IDS sequence $\mathcal{T}_y$, whose global and local textual representations are denoted by $t_{\mathrm{global}}(y)$ and $T_{\mathrm{local}}(y)$, respectively.

First, the global visual representation of the input image is matched with the global textual representations of all candidate categories:
\begin{equation}
S_{\mathrm{global}}(y \mid I)=\langle v_{\mathrm{global}}(I), t_{\mathrm{global}}(y)\rangle, \quad y \in \mathcal{Y}_{\mathrm{all}}
\end{equation}
According to this score, the top-$K$ candidates are selected to form a high-confidence candidate set $\mathcal{Y}_{\mathrm{top}K}$.

Then, local fine-grained interaction is computed only on the Top-$K$ candidates for reranking:
\begin{equation}
S_{\mathrm{local}}(y \mid I)=S_{T2I}(V_{\mathrm{local}}(I), T_{\mathrm{local}}(y)), \quad y \in \mathcal{Y}_{\mathrm{top}K}.
\end{equation}

Finally, the global and local scores are normalized over the Top-$K$ candidate set:
\begin{equation}
P_{\mathrm{global}}(y\mid I)=
\frac{
\exp\left(S_{\mathrm{global}}(y\mid I)/\tau_g\right)
}{
\sum_{y' \in \mathcal{Y}_{\mathrm{top}K}}
\exp\left(S_{\mathrm{global}}(y'\mid I)/\tau_g\right)
}
\end{equation}
\begin{equation}
P_{\mathrm{local}}(y\mid I)=
\frac{
\exp\left(S_{\mathrm{local}}(y\mid I)/\tau_l\right)
}{
\sum_{y' \in \mathcal{Y}_{\mathrm{top}K}}
\exp\left(S_{\mathrm{local}}(y'\mid I)/\tau_l\right)
}
\end{equation}
The final decision score is obtained by multiplicative fusion:
\begin{equation}
S_{\mathrm{final}}(y\mid I)=P_{\mathrm{global}}(y\mid I)\cdot P_{\mathrm{local}}(y\mid I).
\end{equation}

This fusion introduces no extra learnable parameters and combines holistic structural evidence with local component evidence. Through this coarse-to-fine design, GL-HPN restricts expensive local interaction to a small set of high-confidence candidates rather than exhaustively matching all candidates, achieving an effective trade-off between recognition accuracy and inference efficiency.
\section{Experiments}

\subsection{Datasets and Experimental Setup}

We conduct experiments on two handwritten Chinese character datasets, CASIA-HWDB1.0-1.1 and ICDAR2013. CASIA-HWDB1.0-1.1 contains 2,678,424 samples from 3,881 categories, among which 3,755 are first-level commonly used Chinese characters. ICDAR2013 contains 224,419 handwritten character samples from 60 writers and covers the same 3,755 commonly used categories.

Following the standard zero-shot evaluation protocol in CCR-CLIP\cite{yu2023chinese}, we split the 3,755 commonly used Chinese characters into seen and unseen subsets according to the predefined category order used in prior work.For training setting, the first $m$ categories are treated as seen classes, where $m \in \{500,1000,1500,2000,2755\}$. Training samples of these seen classes are taken from CASIA-HWDB1.0-1.1. The last 1000 categories are treated as unseen classes, and their samples in ICDAR2013 are used for evaluation. For the textual side, we use the IDS decomposition dictionary adopted in prior work.

The visual encoder is ResNet-50 and the text encoder is a 12-layer Transformer. Input images are resized to $128 \times 128$, the batch size is 128, and the total number of training epochs is 50. The warm-up stage occupies the first 25\% epochs, after which the global and local loss weights are linearly adjusted with $\alpha=0.8$ and $\beta=1.0$. All experiments are implemented in PyTorch and conducted on a single NVIDIA RTX4090D GPU. Unless otherwise specified, the coarse retrieval stage uses $K=50$. We further analyze the sensitivity of hierarchical inference to different $K$ values in the accuracy-efficiency analysis.

We report Top-1 Accuracy, Recall@$K$, and average inference latency (Latency, ms/img). Top-1 Accuracy measures the final recognition accuracy, Recall@$K$ evaluates whether the ground-truth class is recalled into the top-$K$ candidates during coarse retrieval, and Latency measures the average end-to-end inference time excluding data loading and I/O overhead.

\subsection{Comparison with Existing Methods}

To systematically evaluate the effectiveness of GL-HPN on zero-shot Chinese character recognition, we compare it with several representative methods, including predictive methods and recent retrieval-based approaches such as CCR-CLIP, FT-CLIP, and GRSTR. Table~\ref{tab:main_results} reports the Top-1 accuracy on Test-1000 under different training scales.

\begin{table}[t]
\centering
\caption{Comparison of zero-shot recognition performance (Top-1 Accuracy, \%) under different training scales.}
\label{tab:main_results}
\resizebox{\columnwidth}{!}{
\begin{tabular}{lccccc}
\toprule
Method & 500 & 1000 & 1500 & 2000 & 2755 \\
\midrule
DenseRAN\cite{wang2018denseran} & 1.70 & 8.44 & 14.71 & 19.51 & 30.68 \\
HDE\cite{cao2020zero} & 4.90 & 12.77 & 19.25 & 25.13 & 33.49 \\
ACPM\cite{zu2022chinese} & 9.72 & 18.50 & 27.74 & 34.00 & 42.43 \\
SideNet\cite{li2024sidenet} & 5.10 & 16.20 & 33.80 & 44.10 & 50.30 \\
HierCode\cite{zhang2025hiercode} & 6.22 & 20.71 & 35.39 & 45.67 & 56.21 \\
STAR\cite{zeng2022star} & 7.54 & 19.47 & 27.79 & 35.53 & 43.86 \\
RSST\cite{yu2024chinese} & 11.56 & 21.83 & 35.32 & 39.22 & 47.44 \\
JRED\cite{luo2025joint} & 25.19 & 36.21 & 41.93 & 42.05 & 56.26 \\
CCR-CLIP\cite{yu2023chinese} & 21.79 & 42.99 & 55.86 & 62.99 & 72.98 \\
FT-CLIP\cite{hong2025improving} & 26.17 & 50.68 & 62.25 & 70.39 & 78.96 \\
GRSTR\cite{dong2026graph} & 32.87 & 56.06 & 67.74 & \textbf{73.41} & \textbf{80.62} \\
\midrule
GL-HPN & \textbf{36.19} & \textbf{57.36} & \textbf{68.84} & 73.30 & 80.04 \\
\bottomrule
\end{tabular}
}
\end{table}

As shown in Table~\ref{tab:main_results}, GL-HPN performs particularly well under low- and medium-resource settings. On Train-500, Train-1000, and Train-1500, our method achieves 36.19\%, 57.36\%, and 68.84\% Top-1 accuracy, respectively, outperforming all compared methods. In the most challenging Train-500 setting, GL-HPN improves over FT-CLIP by 10.02 points, over CCR-CLIP by 14.40 points, and over GRSTR by 3.32 points. These results suggest that the proposed global-local modeling improves generalization to unseen characters, especially when only limited seen categories are available.

When the training scale further increases to Train-2000 and Train-2755, GL-HPN reaches 73.30\% and 80.04\%, respectively, remaining highly competitive with the strongest baseline GRSTR. Overall, GL-HPN demonstrates stable performance across different seen-class scales and exhibits more significant advantages under low-resource settings.

\subsection{Ablation Study}

To verify the effect of each module, we conduct ablation studies on the most challenging Train-500 setting, using a FILIP-style local fine-grained alignment model as the baseline. Since the pure global-only setting corresponds to the standard holistic retrieval paradigm already represented by CCR-CLIP in Table~\ref{tab:main_results}, we focus the ablation here on the fine-grained alignment setting in order to isolate the contributions of the proposed dual-branch design and the structure filtering mask. The results are shown in Table~\ref{tab:ablation}.

\begin{table}[t]
\centering
\caption{Ablation study results.}
\label{tab:ablation}
\begin{tabular}{lccc c}
\toprule
Model & Global & Local & SFM & Top-1 Acc \\
\midrule
Baseline (FILIP) & $\times$ & $\checkmark$ & $\times$ & 34.54 \\
GL-HPN (w/o SFM) & $\checkmark$ & $\checkmark$ & $\times$ & 34.19 \\
GL-HPN (Ours) & $\checkmark$ & $\checkmark$ & $\checkmark$ & \textbf{36.19} \\
\bottomrule
\end{tabular}
\end{table}

As shown in Table~\ref{tab:ablation}, simply introducing the dual-branch design without structure filtering does not improve performance and even causes a slight drop, from 34.54\% to 34.19\%. This suggests that, although the dual-branch architecture provides the structural basis for the proposed coarse-to-fine hierarchical inference framework, its effectiveness depends on the stability of the local branch. When structural operators are still involved in local similarity aggregation, their visually unsupported responses may disturb local matching and weaken candidate reranking. After further introducing the structure filtering mask, the accuracy improves from 34.19\% to 36.19\%, outperforming both the dual-branch variant without SFM and the local-only baseline. These results demonstrate that the proposed SFM is necessary for making the global-local framework effective and for stabilizing local fine-grained alignment.

To further understand these observations, we visualize the patch-token response maps in the local branch. For each selected token, we compute its cosine similarity with all $N_p$ visual patch features and reshape the resulting responses into spatial heatmaps over the input image.

\begin{figure}[t]
    \centering
    \includegraphics[width=0.82\columnwidth]{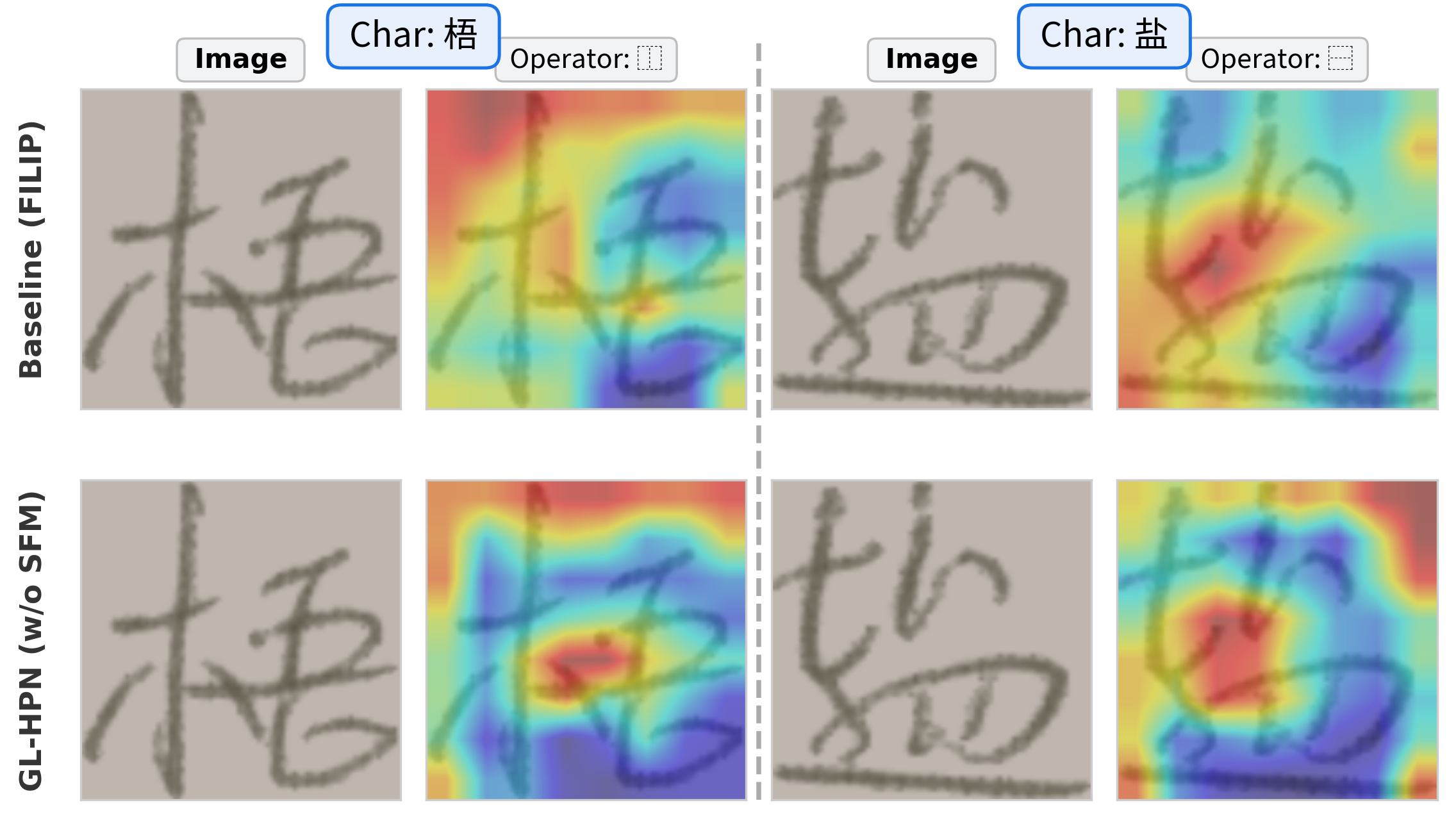}
    \caption{Response visualization of a structural operator token. The operator produces scattered and unstable local responses, indicating that it does not correspond to a concrete visual entity and may introduce noise when directly involved in local similarity aggregation.}
    \label{fig:operator_heatmap}
\end{figure}

Fig.~\ref{fig:operator_heatmap} shows the local response maps of a representative structural operator token. It can be observed that the responses are spatially scattered and do not form a stable localized activation pattern. This suggests that structural operators do not correspond to concrete visual entities in the character image. Therefore, if such tokens are directly included in patch-token matching and local score aggregation, they may inject spurious responses into the local alignment process and interfere with the discriminative contribution of radical tokens.

\begin{figure}[t]
    \centering
    \includegraphics[width=\columnwidth]{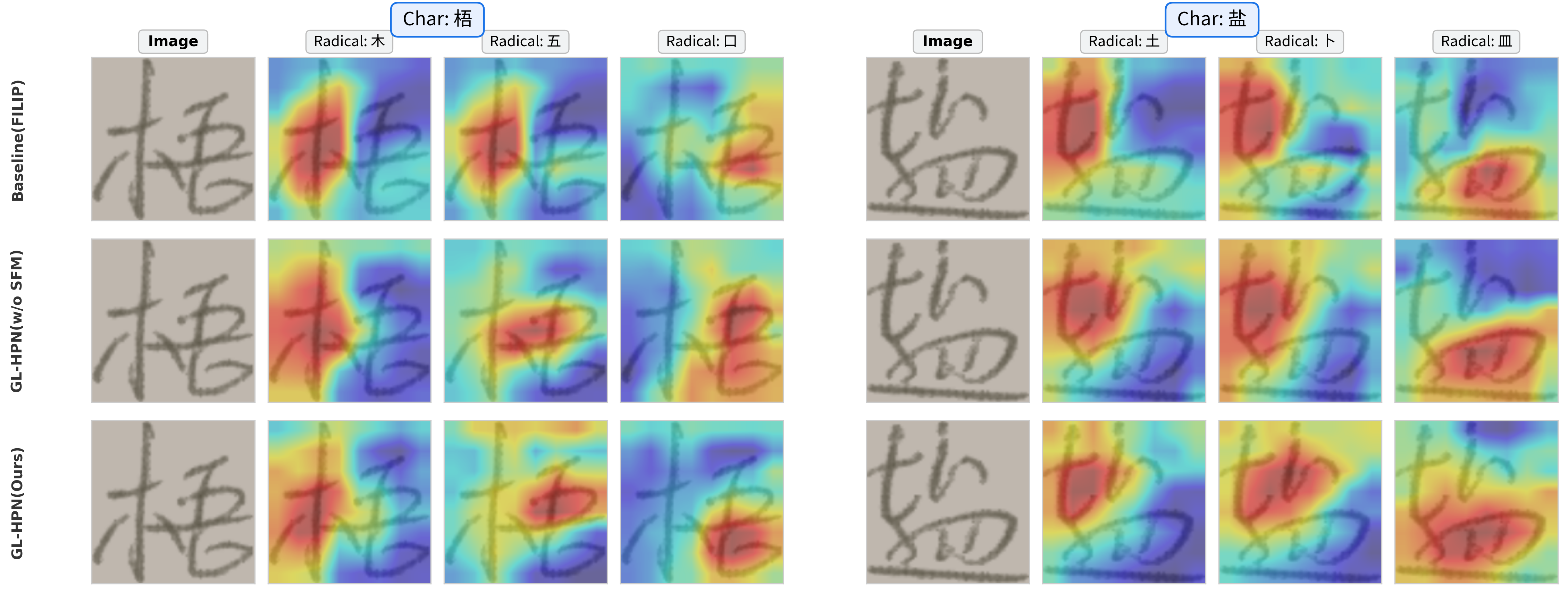}
    \caption{Response visualization of a radical token under different settings. Compared with the FILIP baseline, the dual-branch design produces clearer region-token correspondence, while the additional use of SFM further concentrates the response on the actual component region.}
    \label{fig:radical_heatmap}
\end{figure}

We further compare the response maps of a representative radical token under the three settings in Fig.~\ref{fig:radical_heatmap}. Compared with the FILIP baseline, GL-HPN (w/o SFM) already produces clearer and more localized responses on the corresponding component region, indicating that the global-local dual-branch design strengthens component-aware fine-grained correspondence. However, when structural operators are still involved in local aggregation, their scattered responses may perturb the final local similarity, making the contribution of the local branch less stable in candidate reranking. After introducing SFM, the response of the radical token becomes more concentrated and better aligned with the actual component region, which is consistent with the improvement observed in Table~\ref{tab:ablation}.

Overall, the quantitative and qualitative results are consistent. The dual-branch design improves component-level correspondence and provides the architectural basis for the proposed coarse-to-fine hierarchical inference strategy. At the same time, structural operators may produce spurious local responses because they lack concrete visual grounding. The proposed structure filtering mask alleviates this issue and enables the global-local framework to achieve more reliable local alignment and better recognition accuracy.

\subsection{Accuracy-Efficiency Analysis of Hierarchical Inference}

To evaluate the effectiveness of the proposed coarse-to-fine hierarchical inference strategy, we further compare recall, final accuracy, and inference latency under different candidate sizes $K$. This analysis also serves as a sensitivity study of the hierarchical inference procedure. For fair comparison, both GL-HPN and the baseline model are evaluated under a two-stage retrieval pipeline: the first stage performs candidate recall over the full category set, and the second stage applies fine-grained interaction and reranking only within the Top-$K$ candidates. Since the baseline model does not contain an explicit global branch, we use spatially averaged local patch features as a proxy global representation for the first-stage recall.

\begin{table}[t]
\centering
\caption{Accuracy-efficiency comparison under selected candidate sizes.}
\label{tab:efficiency}
\resizebox{\columnwidth}{!}{
\begin{tabular}{ccccc}
\toprule
$K$ & Method & Recall@$K$ & Top-1 Acc & Lat. (ms) \\
\midrule
\multirow{2}{*}{10}
& GL-HPN   & 63.94 & 36.15 & 1.09 \\
& Baseline & 35.25 & 24.39 & 1.19 \\
\midrule
\multirow{2}{*}{30}
& GL-HPN   & 76.27 & 36.19 & 1.10 \\
& Baseline & 51.60 & 29.76 & 1.22 \\
\midrule
\multirow{2}{*}{50}
& GL-HPN   & 81.12 & 36.19 & 1.11 \\
& Baseline & 60.15 & 31.76 & 1.22 \\
\midrule
\multirow{2}{*}{100}
& GL-HPN   & 86.88 & 36.19 & 1.20 \\
& Baseline & 71.37 & 33.35 & 1.33 \\
\midrule
\multirow{2}{*}{500}
& GL-HPN   & 95.80 & 36.19 & 2.54 \\
& Baseline & 91.64 & 34.43 & 2.68 \\
\midrule
\multirow{2}{*}{3755}
& GL-HPN   & 100.00 & 36.19 & 14.49 \\
& Baseline & 100.00 & 34.54 & 14.46 \\
\bottomrule
\end{tabular}
}
\end{table}

\begin{figure}[t]
    \centering
    \includegraphics[width=\columnwidth]{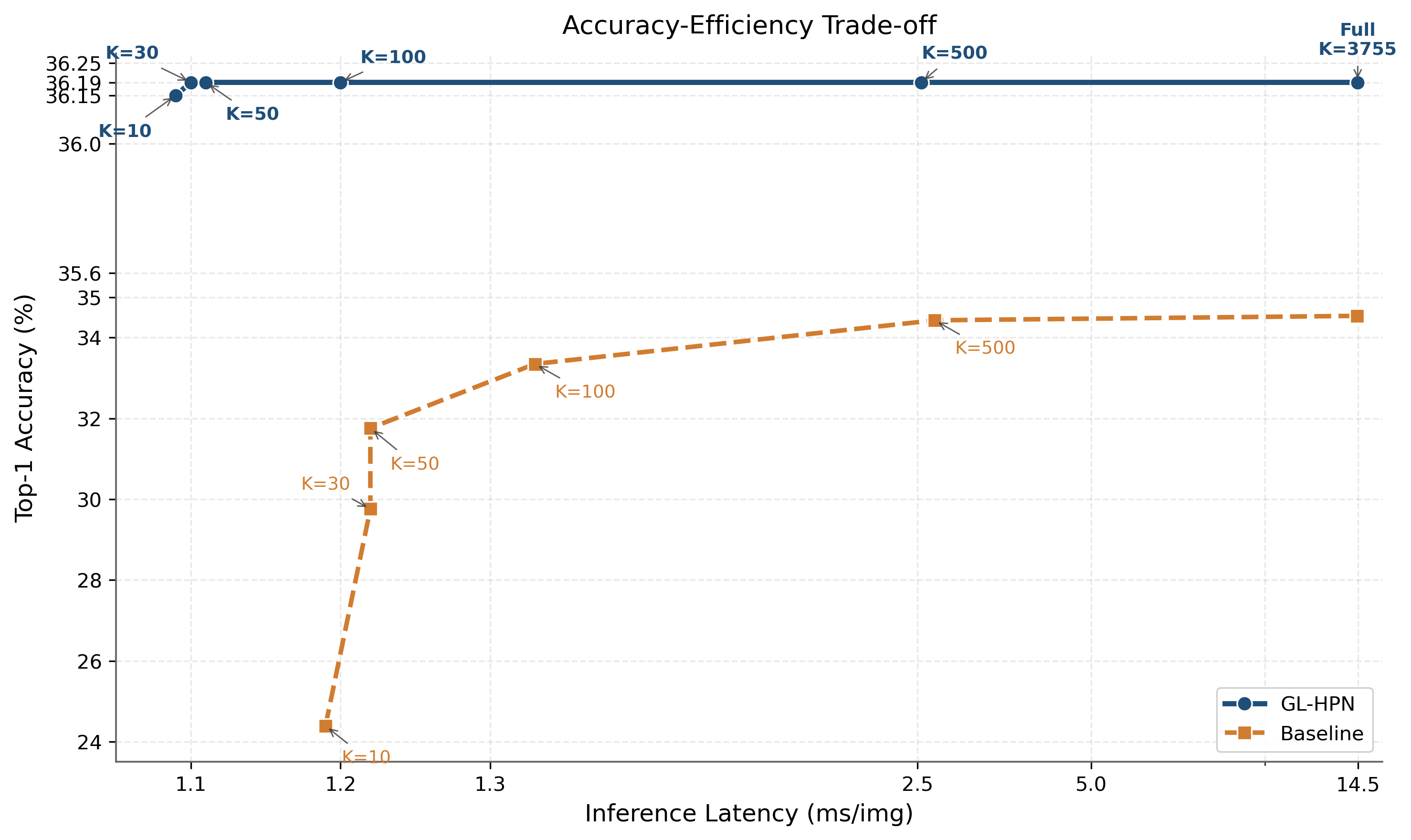}
    \caption{Accuracy-efficiency trade-off under different candidate sizes $K$. For better visualization, the low-latency region on the horizontal axis and the high-accuracy region on the vertical axis are locally expanded. GL-HPN reaches stable recognition accuracy with a relatively small candidate set, while full-candidate local interaction leads to much higher inference latency with limited additional gain.}
    \label{fig:efficiency_tradeoff}
\end{figure}

As shown in Table~\ref{tab:efficiency} and Fig.~\ref{fig:efficiency_tradeoff}, GL-HPN achieves high recall and stable accuracy even with a relatively small candidate size. For example, when $K=30$, Recall@30 already reaches 76.27\%, and Top-1 accuracy reaches 36.19\%, which is identical to the result obtained using the full candidate set, while the average inference latency is only 1.10 ms/img. In contrast, under a similar latency level, the baseline only achieves 29.76\% accuracy and must expand to much larger candidate sets to approach its best performance.

The visualization further shows that GL-HPN reaches its performance plateau at a much smaller candidate size, whereas the baseline continues to rely on larger candidate sets for gradual improvement. This result suggests that the global branch in GL-HPN provides more reliable coarse candidate recall through holistic structural representations. On top of this, the local branch only needs to perform fine-grained discrimination over a small but high-quality candidate set in order to achieve stable recognition performance. Therefore, global-local dual-branch representation learning and coarse-to-fine hierarchical inference are well matched: the former improves recall quality, while the latter substantially reduces the cost of local interaction, together yielding a better trade-off between accuracy and efficiency.

\section{Conclusion}
In this paper, we propose the Global-Local Hierarchical Perception Network (GL-HPN) for zero-shot Chinese character recognition. GL-HPN decouples holistic structural modeling from fine-grained patch-token interaction to learn discriminative representations. By introducing a structure filtering mask (SFM), we successfully mitigate alignment noise caused by non-entity structural operators. Built upon this, our coarse-to-fine hierarchical inference strategy restricts expensive local matching to high-confidence candidates. Experiments demonstrate that GL-HPN achieves an excellent trade-off between accuracy and efficiency, significantly outperforming existing methods in challenging low-resource settings.

{
    \small
    \bibliographystyle{ieeenat_fullname}
    \bibliography{main}
}


\end{document}